\documentclass{article}
\usepackage[a4paper,tmargin=30mm,bmargin=30mm,lmargin=15mm,rmargin=15mm,headsep=20mm]{geometry}

\usepackage{graphicx}
\usepackage{url}
\usepackage{color}

\usepackage{seqsplit}
\usepackage{tcolorbox}
\tcbuselibrary{listingsutf8}
\usepackage{listings}
\usepackage{changepage}
\usepackage{booktabs}
\usepackage{amsmath}
\usepackage{amssymb}
\usepackage{hyperref}
\usepackage{cleveref}
\usepackage{stix}
\usepackage{authblk}

\usepackage{listings}
\usepackage{glossaries}
\newacronym{ai}{AI}{Artificial Intelligence}
\newacronym{eda}{EDA}{Easy Data Augmentation}
\newacronym{ehr}{EHR}{Electronic Health Record}
\newacronym{gdpr}{GDPR}{General Data Protection Regulation}
\newacronym{icd}{ICD-10}{International Classification of Diseases, Tenth Revision}
\newacronym{llm}{LLM}{Large Language Model}
\newacronym{mmd}{MMD}{Maximum Mean Discrepancy}
\newacronym{mse}{MSE}{Mental Status Examination}
\newacronym{nlp}{NLP}{Natural Language Processing}
\newacronym{nnd}{NND}{Nearest Neighbor Distance}
\newacronym{sms}{SMS}{Sentence Mover’s Similarity}
\newacronym{tsne}{t-SNE}{t-distributed Stochastic Neighbor Embedding}
\newacronym{ttr}{TTR}{Type-Token Ratio}
\newacronym{umap}{UMAP}{Uniform Manifold Approximation and Projection}

\definecolor{newblue}{HTML}{9ECCFE}
\definecolor{newgreen}{HTML}{A2FBC1}
\definecolor{newyellow}{HTML}{EFED9A}
\definecolor{newred}{HTML}{F7AC88}

\begin{document}

\title{Fidelity, Diversity, and Privacy: A Multi-Dimensional LLM Evaluation for Clinical Data Augmentation}

\author[1]{Guillermo Iglesias \thanks{guillermo.iglesias@upm.es}}
\author[1]{Gema Bello-Orgaz \thanks{gema.borgaz@upm.es}}
\author[1]{María Navas-Loro \thanks{m.navas@upm.es}}
\author[1]{Cristian Ramirez-Atencia \thanks{cristian.ramirez@upm.es}}
\author[2]{Mercè Salvador Robert \thanks{merce.salvador@salud.madrid.org}}
\author[3, 2, 4, 5, 6, 7, 8, ]{Enrique Baca-Garcia \thanks{enrique.baca@uam.es}}

\affil[1]{Departamento de Sistemas Informáticos, Escuela Técnica Superior de Ingeniería de Sistemas Informáticos, Universidad Politécnica de Madrid, Spain}
\affil[2]{Department of Psychiatry, University Hospital Rey Juan Carlos, Mostoles, Spain}
\affil[3]{Department of Psychiatry, University Hospital Jimenez Diaz Foundation, Madrid, Spain}
\affil[4]{Department of Psychiatry, General Hospital of Villalba, Madrid, Spain}
\affil[5]{Department of Psychiatry, University Hospital Infanta Elena, Valdemoro, Spain}
\affil[6]{Department of Psychology, Universidad Catolica del Maule, Talca, Chile}
\affil[7]{Department of Psychiatry, Madrid Autonomous University, Madrid, Spain}
\affil[8]{CIBERSAM, ISCIII}

\date{}

\maketitle

\begin{abstract}
The scarcity of high-quality annotated medical data, particularly in mental health, poses a significant bottleneck for training robust machine learning models. Privacy regulations restrict data sharing, making synthetic data generation a promising alternative. The use of \glspl{llm} in a data augmentation pipeline could be leveraged as an alternative in this field. In the proposed methodology, \texttt{DeepSeek-R1}, \texttt{OpenBioLLM-Llama3} and \texttt{Qwen 3.5} are used to generate synthetic mental health evaluation reports conditioned on specific \gls{icd} codes. Because naive text generation can lead to mode collapse or privacy breaches (memorization), a comprehensive evaluation framework is introduced. The generated diagnostic texts are assessed across three dimensions: semantic fidelity, lexical diversity, and privacy/plagiarism. The results demonstrate that all models can generate clinically coherent, diverse, and privacy-safe synthetic reports, significantly expanding the available training data for clinical natural language processing tasks without compromising patient confidentiality.
\end{abstract}


\glsresetall
\section{Introduction}
The digitization of healthcare has led to an exponential increase in the volume of \glspl{ehr}, offering an unprecedented opportunity for the application of \gls{ai} and \gls{nlp} in clinical decision support systems. Among clinical narratives, \gls{mse} reports are particularly rich in unstructured data, capturing psychological, cognitive, and behavioral assessments of patients. However, the development of robust \gls{ai} models for psychiatric and psychological domains is severely hindered by data scarcity. Strict data privacy regulations, such as \gls{gdpr} in Europe, and the highly sensitive nature of mental health records limit the availability, sharing, and utilization of real patient data.

To overcome the barrier of data access, synthetic data generation through Data Augmentation has emerged as a highly promising paradigm. Recent advancements in \glspl{llm} have demonstrated remarkable capabilities in generating coherent, domain-specific text. Despite this potential, generating synthetic medical reports that are both clinically accurate and sufficiently diverse remains a complex challenge. Synthetic clinical text must faithfully reflect the semantic and linguistic properties of real reports associated with specific diagnoses, e.g. \gls{icd} codes\cite{who2016icd10}, while strictly avoiding the memorization of original patient data, which would constitute a privacy violation. Furthermore, models often suffer from mode collapse, producing repetitive text that lacks the natural variability of human-authored clinical notes.

In this paper, a robust methodology for the targeted data augmentation of mental status evaluation reports using state-of-the-art \glspl{llm} is presented. Specifically, we leverage and compare the performance of \texttt{\seqsplit{DeepSeek-R1}}\cite{guo2025deepseek}, \texttt{\seqsplit{OpenBioLLM-Llama3}}\cite{OpenBioLLMs}, and \texttt{\seqsplit{Qwen 3.5}}\cite{qwen3.5} in generating synthetic clinical narratives conditioned on targeted \gls{icd} codes. Using a proprietary database of real medical evaluations, the approach employs a few-shot prompting strategy. By providing the models with a context window of $n=10$ reference cases for a specific \gls{icd} code, the \glspl{llm} are instructed to synthesize $10$ novel, independent reports that maintain diagnostic coherence with the underlying medical condition. The result is a set of 940 new synthetic reports.

A primary contribution of this work is the development and application of a comprehensive evaluation framework designed to rigorously assess the quality of the synthetic medical data. Recognizing that standard generation metrics are insufficient for clinical \gls{nlp}, the generated reports are evaluated across three fundamental axes: domain fidelity (how closely the synthetic text resembles the real distribution), lexical diversity (the variety and richness of the generated language), and plagiarism percentage (based on the average distance between the original text and the generated synthetic text). This evaluation scheme analyzes both raw plain text and dense semantic embeddings extracted via the \texttt{all-MiniLM-L6-v2} model.

To achieve a holistic assessment, a robust battery of metrics are calculated, including semantic similarity measures (\gls{mmd}\cite{gretton2012kernel}, BERTScore\cite{zhang2019bertscore}, \gls{sms}\cite{clark2019sentence}, \gls{nnd}\cite{clark1954distance}), standard translation and n-gram overlap metrics (ROUGE-1, ROUGE-2, ROUGE-L\cite{lin2004rouge}, METEOR\cite{banerjee2005meteor}) and diversity indicators (Self-BLEU\cite{zhu2018texygen}, \gls{ttr}\cite{templin1957certain}). Furthermore, this quantitative analysis is complemented by empirical evaluations, specifically through the visual comparison of the original and synthetic embedding spaces using dimensionality reduction techniques (\gls{umap}\cite{mcinnes2018umap} and \gls{tsne}\cite{van2008visualizing}), alongside frequent n-gram distribution analysis.

The remainder of this paper is structured as follows: \Cref{section:RelatedWork} reviews related work in clinical synthetic data generation. \Cref{section:Methodology} details the methodology, including the data description, the \gls{llm} prompting strategy, the augmentation pipeline and the evaluation framework. \Cref{section:ResultsDiscussion} discusses the results and comparative performance of the models. \Cref{section:Conclusion} contains the conclusions of the research.

\section{Related Work}
\label{section:RelatedWork}
The challenge posed by the scarcity of annotated text datasets in \gls{nlp} has given rise to a diverse ecosystem of data augmentation techniques designed to expand training datasets without the need for manual labeling. Early strategies focused primarily on rule-based manipulations (such as the \gls{eda} framework \cite{wei2019eda}) or neural network-based approaches (applying retranslation \cite{shorten2021text} or “Mixup” interpolation \cite{guo2021augmenting}). Although these methods are effective for generating general synthetic text,  they often lack the contextual nuances required for specific applications in particular domains.

In the clinical setting, the generation of synthetic \gls{ehr} data is further complicated by \gls{gdpr} regulations and the clinical need to maintain diagnostic consistency. Unlike general text, clinical narratives must adhere to a structured medical logic; therefore, specialized approaches in this specific field have incorporated medical ontologies to ensure that the generated symptoms are accurate \cite{spasic2020clinical}. This requirement for accuracy has led to research on prompt engineering. For example, in the work by Moral-González et al. \cite{del2025comparative}, they demonstrated that providing detailed instructions,ranging from basic descriptions to comprehensive clinical guidelines, to generate texts directly influences a model’s ability to maintain medical accuracy. Furthermore, hybrid frameworks have emerged that combine the generative power of \glspl{llm} with traditional rule-based systems to address complex challenges such as the classification of rare diseases \cite{lyu2023hybrid}.

The most recent research focuses on state-of-the-art reasoning models, which have a superior ability to capture the complex reasoning found in a \gls{mse}. Frameworks such as \texttt{AugGPT} have demonstrated the potential of \gls{ai} to reformulate clinical sets with high semantic fidelity \cite{dai2025auggpt}, and other recent studies highlight how \glspl{llm} are particularly well-suited for tasks requiring significant reasoning effort, where the context between symptoms is as important as the symptoms themselves \cite{chai2026text}.

Building on this prior work, our study presents a specific methodology for improving the generation of synthetic text related to clinical reports by leveraging and comparing the performance of different \glspl{llm}. Unlike previous hybrid approaches or simple reformulation, we employ a few-shot prompt strategy strictly governed by \gls{icd} codes to ensure that the synthetic reports maintain high diagnostic consistency. Furthermore, a key contribution of this research is the proposal of a multidimensional evaluation framework for the generated texts, based on various metrics from the field of \gls{nlp}. By integrating lexical, semantic, and plagiarism indicators into this evaluation framework, we provide validation to ensure that the generated data are both clinically viable and privacy-safe.

\section{Methodology}
\label{section:Methodology}
\subsection{Clinical Dataset}
The foundation of this research is a proprietary clinical dataset comprising $25,803$ real medical reports. These records were systematically extracted from the psychiatric and psychological emergency clinic of Fundación Jiménez Díaz (Grupo Quirón Salud - Madrid, Spain). Using data from a real-world emergency clinical setting is highly advantageous, as it faithfully captures the linguistic nuances, urgency, and unstructured nature typical of frontline mental health assessments.

Each record in the dataset contains a free-text \gls{mse} narrative detailing the patient's diagnostic evaluation. Given the highly descriptive nature of psychiatric assessments, the reports exhibit significant variation in length and detail. On average, text lengths are $48.3$ characters, with a minimum of $2$ and a maximum of $2,560$ characters, reflecting a highly heterogeneous corpus that ranges from concise emergency summaries to extensive clinical histories.

Crucially for the augmentation task, each report is annotated with one or more diagnostic codes corresponding to the \gls{icd}. The dataset is formulated as a multi-label classification problem, where a single report may be assigned multiple concurrent diagnoses, capturing clinical comorbidities.

In total, the dataset encompasses a label space of $94$ distinct \gls{icd} categories. An analysis of the label distribution of the dataset reveals a strong prevalence of the \gls{icd} 'F' chapter, specifically Mental, Behavioral and Neurodevelopmental disorders. This includes, but is not limited to, mood (affective) disorders (e.g., F30-F39 such as Bipolar disorders and Depressive episodes), anxiety, dissociative, stress-related, and somatoform disorders (e.g., F40-F48), as well as mental and behavioral disorders due to psychoactive substance use (e.g., F10-F19). The presence of these 94 specific diagnostic labels serves as the conditioning ground truth for the targeted synthetic data generation process performed by the \glspl{llm}.

\subsection{Prompt Engineering and LLM Setup}
To carry out the targeted data augmentation, we evaluated three state-of-the-art \glspl{llm}: \texttt{\seqsplit{DeepSeek-r1:32b}}, \texttt{\seqsplit{OpenBioLLM-Llama3-70b.i1-q4\_k\_m.gguf}}, and \texttt{\seqsplit{Qwen 3.5:35b-a3b-q8\_0}}. The selection of these specific medium-sized models, alongside the use of quantized versions (such as the 4-bit and 8-bit quantizations for \texttt{OpenBioLLM-Llama3} and \texttt{Qwen3.5}, respectively), is a strategic decision to balance advanced natural language understanding and clinical domain knowledge with the constraints of privacy concerns in the medical field.

Our generative approach relies on a meticulously designed few-shot prompting strategy to instruct the models to synthesize realistic, medically accurate Spanish clinical narratives. By providing in-context learning examples, the models are guided to emulate the specific linguistic style, length, and terminology of the original emergency department reports.

The prompt architecture is modularly divided into a system prompt and a structured user query, as detailed below.

\textbf{System Prompt}: The models are initially contextualized through a system prompt that defines their persona and output constraints. The \gls{llm} is instructed to act as a specialized medical assistant tasked with generating brief clinical diagnoses in Spanish. Crucially, to facilitate automated parsing in the data pipeline, the system prompt enforces a strict structural constraint, mandating that the output must be exclusively a valid, correctly closed JSON object without any conversational filler or markdown formatting.
\newline\newline
\noindent
\begin{minipage}{0.49\textwidth}
    \begin{lstlisting}[
        frame=single,
        title={system\_prompt},
        xleftmargin=2mm,
        xrightmargin=2mm,
        basicstyle=\ttfamily\scriptsize,
        breaklines=true,
        breakindent=0pt
    ]
Eres un asistente medico especializado en generar diagnosticos clinicos breves en espanol.
Devuelve unicamente un JSON **valido y cerrado correctamente** (todas las llaves deben estar balanceadas).
No incluyas comillas triples ni texto adicional.

El JSON debe tener esta estructura exacta: 
{
    "diagnosticos": ["Diagnostico A", "Diagnostico B", ...]
}
    \end{lstlisting}
\end{minipage}
\hfill
\begin{minipage}{0.49\textwidth}
    \begin{lstlisting}[
        frame=single,
        title={system\_prompt (English translation)},
        xleftmargin=2mm,
        xrightmargin=2mm,
        basicstyle=\ttfamily\scriptsize,
        breaklines=true,
        breakindent=0pt
    ]
You are a medical assistant specialized in generating brief clinical diagnoses in Spanish.
Return only a **valid and properly closed** JSON (all curly braces must be balanced).
Do not include triple quotes or additional text.

The JSON must have this exact structure: 
{
    "diagnosis": ["Diagnosis A", "Diagnosis B", ...]
}
    \end{lstlisting}
\end{minipage}
\newline\newline

\textbf{User Query Structure:} The main prompt delivered to the model is dynamically constructed for each target diagnostic label using three main components:

\begin{enumerate}
    \item \textbf{Context Initialization} (\texttt{start\_prompt}): This introductory segment informs the model that it will be provided with an \gls{icd} code and a set of associated real clinical examples.
    \newline\newline
    \noindent
    \begin{minipage}{0.49\textwidth}
        \begin{lstlisting}[
            frame=single,
            title={start\_prompt},
            xleftmargin=2mm,
            xrightmargin=2mm,
            basicstyle=\ttfamily\scriptsize,
            breaklines=true,
            breakindent=0pt
        ]
    A continuacion tienes informacion sobre una etiqueta CIE-10 y ejemplos de diagnosticos asociados:
        \end{lstlisting}
    \end{minipage}
    \hfill
    \begin{minipage}{0.49\textwidth}
        \begin{lstlisting}[
            frame=single,
            title={start\_prompt (English translation)},
            xleftmargin=2mm,
            xrightmargin=2mm,
            basicstyle=\ttfamily\scriptsize,
            breaklines=true,
            breakindent=0pt
        ]
    Below is information about a CIE-10 code and examples of associated diagnoses:
        \end{lstlisting}
    \end{minipage}
    \newline\newline

    \item \textbf{Few-Shot Examples} (\texttt{example\_prompt}): For each augmentation cycle, the prompt iterates over $n=10$ real reference cases extracted from our proprietary database. Each example injected into the prompt specifies the \gls{icd} codes, the descriptive name of the label, and the verbatim text of the clinical report. This grounds the model in the actual distribution of the data.
    \newline\newline
    \noindent
    \begin{minipage}{0.49\textwidth}
        \begin{lstlisting}[
            frame=single,
            title={example\_prompt},
            xleftmargin=2mm,
            xrightmargin=2mm,
            basicstyle=\ttfamily\scriptsize,
            breaklines=true,
            breakindent=0pt
        ]
    - Codigos CIE-10 asociados al ejemplo: {code}
    - Nombre de la etiqueta: {code_name}
    - Diagnostico de ejemplo: {example}
        \end{lstlisting}
    \end{minipage}
    \hfill
    \begin{minipage}{0.49\textwidth}
        \begin{lstlisting}[
            frame=single,
            title={example\_prompt (English translation)},
            xleftmargin=2mm,
            xrightmargin=2mm,
            basicstyle=\ttfamily\scriptsize,
            breaklines=true,
            breakindent=0pt
        ]
    - ICD-10 codes associated with the example: {code}
    - Label name: {code_name}
    - Example diagnosis: {example}
        \end{lstlisting}
    \end{minipage}
    \newline\newline

    where \texttt{code} is the corresponding \gls{icd} code, \texttt{code\_name} is the description of the code and \texttt{example} is one random example from the real dataset.
    
    \item \textbf{Task Formulation} (\texttt{end\_prompt}): The final segment synthesizes the objective. It explicitly instructs the model to generate a precise number of new, distinct clinical diagnoses corresponding to the target \gls{icd} code. It reiterates the strict JSON formatting requirement, mapping each new case to its generated diagnostic text.
    \newline\newline
    \noindent
    \begin{minipage}{0.49\textwidth}
    \begin{lstlisting}[
        frame=single,
        title={end\_prompt},
        xleftmargin=2mm,
        xrightmargin=2mm,
        basicstyle=\ttfamily\scriptsize,
        breaklines=true,
        breakindent=0pt
    ]
    Tu tarea es generar exactamente {n} diagnosticos clinicos nuevos, distintos de los anteriores, que correspondan a la etiqueta {code}-{code_name}.
    
    Devuelve la respuesta como un JSON con esta estructura exacta:
    
    {
      "caso 1": {
          "[lista_codigos]":"Diagnostico 1"
      },
      "caso 2": {
          "[lista_codigos]":"Diagnostico 2"
      },
      ...
    }
    
    Debe ser JSON valido, sin ningun texto adicional.
        \end{lstlisting}
    \end{minipage}
    \hfill
    \begin{minipage}{0.49\textwidth}
        \begin{lstlisting}[
            frame=single,
            title={end\_prompt (English translation)},
            xleftmargin=2mm,
            xrightmargin=2mm,
            basicstyle=\ttfamily\scriptsize,
            breaklines=true,
            breakindent=0pt
        ]
    Your task is to generate exactly {n} new clinical diagnoses, distinct from the previous ones, that correspond to the label {code}-{code_name}.
    
    Return the response as JSON with this exact structure:
    
    {
      "case 1": {
          "[code_list]":"Diagnosis 1"
      },
      "case 2": {
          "[code_list]":"Diagnosis 2"
      },
      ...
    }
    
    It must be valid JSON, with no additional text.
        \end{lstlisting}
    \end{minipage}
    \newline\newline

    In the \textit{end\_prompt}, \texttt{n} is the number of diagnosis to generate in each \gls{llm} call, \texttt{code} is the corresponding \gls{icd} code to generate and \texttt{code\_name} is the description of the code.
\end{enumerate}

By structuring the input in this modular fashion, we ensure that the \glspl{llm} are deeply contextualized by real diagnostic nuances while being heavily constrained in their output format, thereby maximizing the efficiency and reliability of the automated synthetic data generation pipeline.

\subsection{Inference and Generation Pipeline}
To generate synthetic data, we implemented an automated iterative processing workflow. The inference process is systematically structured to cover the entire spectrum of \gls{icd} codes present in the dataset.

For each \gls{icd} code processed, the generation algorithm selects $m=10$ real clinical cases that have been previously reviewed and validated. These $10$ cases are dynamically integrated into the prompt via the \texttt{example\_prompt} component, establishing a robust few-shot learning paradigm. By providing this number of examples, the model has sufficient lexical and structural variability to understand the nuances of the specific disorder without exceeding the context window.

After presenting the examples, an inference call is made to the \gls{llm}, with the instruction to generate $n=10$ new and unique synthetic diagnoses for a target code. This \textit{10 input examples for 10 output diagnoses} loop is repeated for each \gls{icd} code and for each of the models (\texttt{DeepSeek-R1}, \texttt{OpenBioLLM\-Llama3}, and \texttt{Qwen 3.5}).

Since \glspl{llm} can generate additional conversational text (\textit{format hallucinations}) despite the instructions in the system prompt, the pipeline includes a critical post-processing step. The raw response returned by the model is analyzed using regular expressions to locate and extract only the block of text enclosed in curly braces, ensuring that the underlying JSON object is captured.

\subsection{Evaluation Framework}
To assess the quality, utility, and safety of the generated synthetic clinical reports, we implemented a multi-dimensional evaluation scheme. Relying on standard n-gram overlap metrics is often inadequate for clinical \gls{nlp}, as medical text requires adherence to domain knowledge. Thus, our evaluation framework is divided into three core dimensions: Semantic Fidelity, Lexical Diversity and Novelty, and Privacy and Plagiarism Analysis. Both raw plain texts and dense semantic embeddings, extracted via the \texttt{all-MiniLM-L6-v2} sentence transformer, were utilized to compute the metrics.

\subsubsection{Semantic Fidelity}
This dimension evaluates how closely the synthetic diagnoses resemble the real clinical reports in terms of meaning, contextual structure, and overall domain distribution.

\begin{itemize}
    \item \textbf{\acrfull{mmd}\cite{gretton2012kernel}:} This statistical measure evaluates domain fidelity by calculating the distance between the embedding distributions of the real and synthetic reports. A lower \gls{mmd} indicates that the \gls{llm} has successfully approximated the true data manifold of the original psychiatric emergency records.

    \item \textbf{BERTScore\cite{zhang2019bertscore}:} This metric leverages pre-trained language models to compute token-level semantic similarity (reporting Precision, Recall, and F1-score) between generated texts and reference cases. It is particularly valuable for capturing deep contextual equivalence rather than relying on exact word matches.

    \item \textbf{\gls{sms}\cite{clark2019sentence}:} Adapting the concept of Word Mover's Distance, \gls{sms} evaluates the continuous semantic representation at the sentence level, assessing the distance required to align the synthetic text's meaning with the original clinical reports.

    \item \textbf{N-gram Overlap Metrics (ROUGE\cite{lin2004rouge} \& METEOR\cite{banerjee2005meteor}):} We compute standard machine translation metrics (ROUGE-1, ROUGE-2, ROUGE-L, and METEOR) to measure surface-level linguistic alignment. These metrics quantify how well standard clinical phrasing, terminology, and syntax are preserved in the augmented data.
\end{itemize}
    Additionally, an empirical visual assessment was carried out. To qualitatively validate the distributional alignment, we apply dimensionality reduction algorithms, specifically \gls{umap}\cite{mcinnes2018umap} and \gls{tsne}\cite{van2008visualizing} to project the high-dimensional \texttt{all-MiniLM-L6-v2} embeddings into a 2D space. This visual comparison provides immediate insight into the structural overlap between the real and synthetic cohorts.

\subsubsection{Lexical Diversity}
A common failure mode in generative models, particularly when conditioned on specific prompts, is \textit{mode collapse}, where the model produces highly repetitive, template-like text. This subsection measures the linguistic richness and natural variance of the augmented corpus.

\begin{itemize}
    \item \textbf{Self-BLEU\cite{zhu2018texygen}:} BLEU score measures internal diversity of each generated synthetic report against the rest of the generated set. A lower Self-BLEU score is highly desirable, as it indicates minimal internal repetition and confirms that the model is generating novel phrasing rather than defaulting to a standard output.

    \item \textbf{\acrfull{ttr}\cite{templin1957certain}:}  measures the vocabulary richness of the synthetic corpus by calculating the ratio of unique words (types) to the total number of words (tokens). A higher \gls{ttr} reflects a varied, expansive, and natural clinical lexicon.
\end{itemize}

\subsubsection{Privacy/Plagiarism Preservation}
Given the highly sensitive nature of \gls{mse} reports, generating synthetic data must strictly avoid memorizing and regurgitating the few-shot reference examples. Such memorization would constitute a critical privacy breach, defeating the purpose of synthetic data generation.

\begin{itemize}
    \item \textbf{\gls{nnd}\cite{clark1954distance}:} For every synthetic report, we calculate the cosine distance to its closest real report in the semantic embedding space. While the average \gls{nnd} serves as a proxy for structural similarity. The greater this distance, the more it indicates that the generation of new data that is almost identical to the original is being avoided. 

    \item \textbf{Plagiarism Rate:} This metric measures the proportion of synthetic reports that are excessively similar to real reports, thereby indicating potential data leakage. It is computed as the percentage of synthetic texts whose \gls{nnd} to a real patient record falls below a predefined distance threshold in the normalized embedding space. A synthetic report is considered plagiarized if its \gls{nnd} satisfies $d < 0.05$.
    
\end{itemize}

\subsection{Experimental setup}
\label{section:ExpSetup}
Experiments and tests were conducted on two 48 GB Nvidia Quadro RTX 8000 GPUs and an Intel Xeon Bronze 3206R CPU @ 1.90GHz. To download and infer data with \glspl{llm} the Ollama package version 0.17.7 was used. All source code is open and accessible at \url{https://------}\footnote{\textit{Code repository will be available at article's acceptance}}.

The experiments performed for data augmentation inference took 4 hours, 30 minutes and 21 seconds, and the evaluation of the results required 27 minutes and 9 seconds.

\section{Results and Discussion}
\label{section:ResultsDiscussion}
To evaluate the efficacy of the proposed data augmentation pipeline, we conducted a comparative analysis of the synthetic clinical records generated by \texttt{DeepSeek-R1}, \texttt{OpenBioLLM-Llama3}, and \texttt{Qwen 3.5}. \Cref{tab:da_results} summarizes the performance of each model across our predefined metrics, encompassing semantic fidelity, lexical diversity, and privacy/plagiarism preservation.

\begin{table}[h!]
    \begin{tabular}{l|cccccc} 
    \toprule
    Model Name &
    \textbf{\gls{mmd} $\blacktriangledown$} &
    \textbf{$\overline{\text{BERTScore F1}} \ \blacktriangle$} &
    \textbf{$\overline{\text{\gls{sms}}} \ \blacktriangledown$} &
    \textbf{Rouge-1 $\blacktriangledown$} &
    \textbf{Rouge-2 $\blacktriangledown$} &
    \textbf{Rouge-L $\blacktriangledown$} \\
    \midrule
    \texttt{DeepSeek-R1:32b} & \textbf{0.012} & \textbf{0.690} & 0.514 &  0.074 & 0.005 & 0.068 \\
    \texttt{OpenBioLLM-Llama3:70b} & 0.027 & 0.686 & \textbf{0.431} & 0.064 & 0.007 & 0.059 \\
    \texttt{Qwen 3.5:35b} & 0.015 & 0.689 & 0.500 & \textbf{0.059} & \textbf{0.003} & \textbf{0.057} \\ 
    \bottomrule
    Dimension & \multicolumn{5}{c}{Semantic fidelity} \\
    \hline
    \end{tabular}

    \vspace{.75em}

    \begin{tabular}{l|c|cc|cc} 
    \toprule
    Model Name &
    \textbf{METEOR $\blacktriangledown$} &
    \textbf{Self BLEU $\blacktriangledown$} &
    \textbf{\gls{ttr} $\blacktriangle$} &
    \textbf{$\overline{\text{\gls{nnd}}} \ \blacktriangle$} &
    \textbf{Plagiarism Rate $\blacktriangledown$} \\
    \midrule
    \texttt{DeepSeek-R1:32b} & 0.056  & 0.134 & 0.986 & 0.186 & 0.060 \\
    \texttt{OpenBioLLM-Llama3:70b} & \textbf{0.040} & 0.234 & 0.978 & 0.203 & 0.100 \\
    \texttt{Qwen 3.5:35b} & 0.053  & \textbf{0.120} & \textbf{0.991} & \textbf{0.232} & \textbf{0.023} \\ 
    \bottomrule
    Dimension & \multicolumn{1}{c|}{Semantic fidelity} & \multicolumn{2}{c|}{Lexical diversity} & \multicolumn{2}{c}{Privacy/Plagiarism preservation}\\
    \cline{1-6}
    \end{tabular}
    \vspace{.5em}
    \caption{Synthetic text metrics for \texttt{DeepSeek-R1}, \texttt{OpenBioLLM-Llama3} and \texttt{Qwen 3.5}. $\blacktriangle$ denotes that the higher the metric, the better; whereas $\blacktriangledown$ denotes that the lower the metric, the better.}
    \label{tab:da_results}
\end{table}

\subsection{Semantic Fidelity}
In terms of overall domain adaptation, \gls{mmd} results reveal that \texttt{DeepSeek-R1} achieved the lowest divergence (0.012) between the synthetic and real distributions. This indicates a superior ability to capture the underlying data manifold of real psychiatric reports.

When evaluating deep semantic equivalence at the sentence and token levels, the Average BERTScore F1 and Average \gls{sms} provided consistent insights. \texttt{DeepSeek-R1} also yielded the highest Average BERTScore F1 of 0.690 whereas  \texttt{OpenBioLLM-Llama3} got the best Average \gls{sms} results with 0.431, demonstrating that the generated narratives not only matched the clinical vocabulary but also maintained the nuanced contextual meaning of the targeted \gls{icd} diagnoses.

Surface-level linguistic overlap was assessed using ROUGE (1, 2, and L) and METEOR metrics. While these metrics penalize deviations from exact phrasing, which is somewhat expected and even desired in synthetic generation to ensure novelty, the results were distributed across models. \texttt{Qwen 3.5:35b} demonstrated the highest novelty in terms of ROUGE, achieving a ROUGE-L of 0.057, while \texttt{OpenBioLLM\-Llama3:70b} maintained the highest baseline structural fidelity with a METEOR score of 0.040.

That said, all the models have similar performance in semantic fidelity, suggesting a high similarity between the original and the synthetic texts.

\begin{figure}[h!]
\begin{adjustwidth}{0cm}{0cm}
    \centering
    \includegraphics[width=\linewidth]{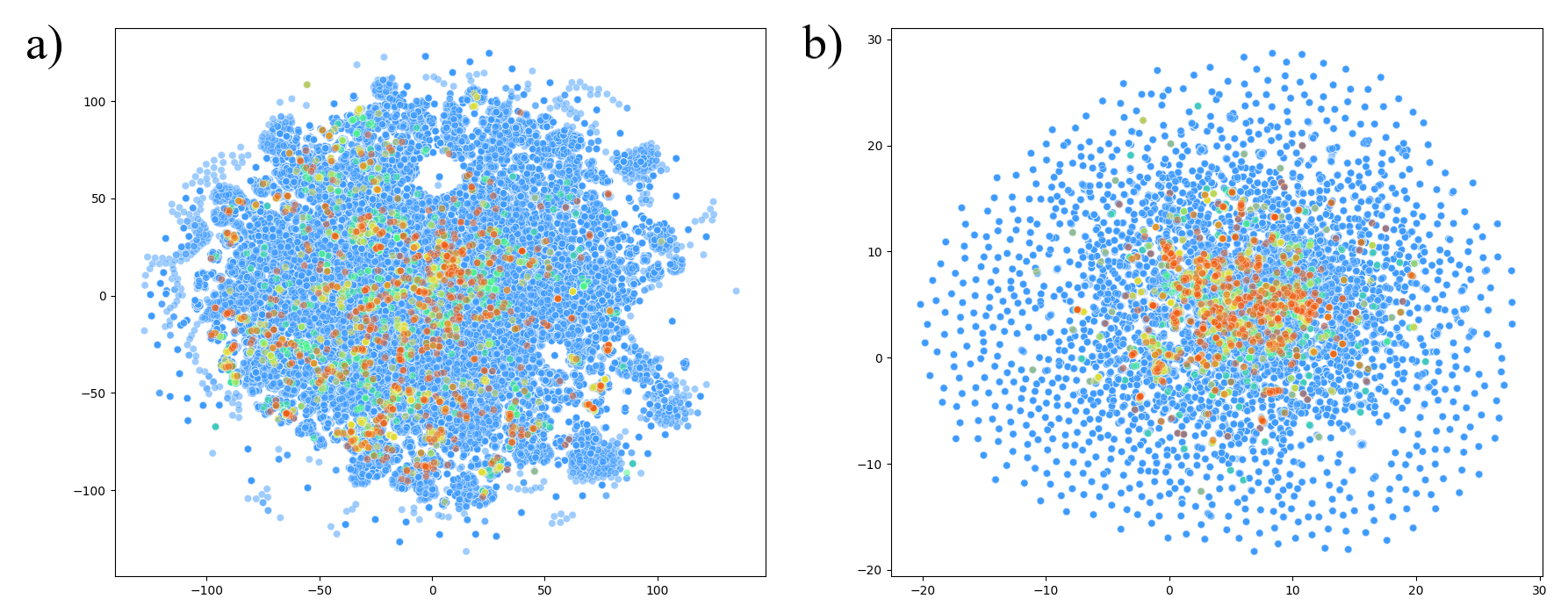}
    \caption{Latent representation of real and synthetic texts using the embeddings produced with \texttt{all-MiniLM-L6-v2} sentence transformer. Dimensionality reduction was performed using a) \gls{tsne} and b) \gls{umap}. \textcolor{newblue}{$\mdsmblkcircle$} shows real data, \textcolor{newgreen}{$\mdsmblkcircle$} shows synthetic data generated with \texttt{DeepSeek-R1}, \textcolor{newyellow}{$\mdsmblkcircle$} shows synthetic data generated with \texttt{OpenBioLLM-Llama3} and \textcolor{newred}{$\mdsmblkcircle$} shows synthetic data generated with \texttt{Qwen 3.5}.}
    \label{fig:VisualAssessment}
\end{adjustwidth}
\end{figure}

\Cref{fig:VisualAssessment} shows an empirical visual assessment using \gls{tsne} and \gls{umap} over the text embeddings. As can be seen, the probability distribution of real and synthetic data is the same, suggesting that there are no visible differences between both sources of data.

\subsection{Lexical Diversity}
A critical success factor in clinical data augmentation is the generation of a heterogeneous corpus. Our analysis of \gls{ttr} and Self-BLEU highlights significant differences in how the models handled vocabulary variance. \texttt{Qwen 3.5} demonstrated the highest vocabulary richness with a \gls{ttr} of 0.991, producing a highly varied lexicon of medical terminology. The other models have similar performance in terms of \gls{ttr}.

\texttt{Qwen 3.5} successfully mitigated mode collapse, recording the lowest Self-BLEU score (0.120). A low Self-BLEU confirms that the model generated distinct, independent reports for each iteration rather than converging on a single, repetitive diagnostic template. In contrast, \texttt{OpenBioLLM-Llama3} exhibited a higher Self-BLEU (0.234), indicating a slight tendency toward structural repetition across its generated outputs.

\subsection{Privacy/Plagiarism Preservation}
To ensure that the few-shot prompting did not result in the memorization and regurgitation of real patient data, we analyzed the \gls{nnd} in the embedding space. All three models maintained an Average \gls{nnd} well above our strict plagiarism threshold ($d < 0.05$). Specifically, \texttt{Qwen 3.5} exhibited an Average \gls{nnd} of 0.232 and the lowest percentage ($2,3\%$) of cases with $d < 0.05$. The distances achieved confirm that while the models learned the semantic distribution of the specific \gls{icd} codes, they generated genuinely synthetic and novel patient histories, thereby successfully preserving patient privacy and passing the data leakage audit.

\section{Conclusion}
\label{section:Conclusion}
The scarcity of high-quality, annotated clinical data, with strict privacy regulations, remains a fundamental bottleneck in the development of \gls{nlp} systems for mental health. In this paper, we proposed and evaluated a comprehensive data augmentation methodology utilizing state-of-the-art \glspl{llm} (\texttt{DeepSeek-R1}, \texttt{OpenBioLLM-Llama3}, and \texttt{Qwen 3.5}) to generate synthetic \gls{mse} reports conditioned on specific \gls{icd} codes.

To overcome the limitations of standard text generation metrics, we introduced a rigorous, multi-dimensional evaluation framework that assesses the generated data across semantic fidelity, lexical diversity, and privacy preservation. Our comparative analysis yielded promising results: all evaluated models demonstrated a strong capacity to approximate the semantic distribution of real clinical records. Specifically, \texttt{DeepSeek-R1} showed the highest domain semantic fidelity, achieving the lowest \gls{mmd} (0.012) and the highest BERTScore F1 (0.690). Conversely, \texttt{OpenBioLLM-Llama3} provided the best alignment in continuous sentence representation (Average \gls{sms} of 0.431) and the lowest surface-level overlap (METEOR of 0.040), although it exhibited slightly higher ROUGE scores compared to \texttt{Qwen 3.5}. On the other hand, \texttt{Qwen 3.5} excelled in producing the most lexically diverse text, successfully mitigating mode collapse with the lowest Self-BLEU score (0.120) and the highest \gls{ttr} (0.991).

Crucially, our privacy auditing confirmed that the applied few-shot prompting strategy did not trigger the memorization of original patient data. All models maintained an Average \gls{nnd} well above the strict plagiarism threshold ($d < 0.05$), ensuring that the synthetic narratives are genuinely novel and safely preserve patient confidentiality.

These findings demonstrate that targeted data augmentation via medium-sized, quantized \glspl{llm} is a highly viable, secure, and efficient solution for expanding clinical data. Future work will focus on integrating the synthetic corpus into downstream clinical classification tasks to quantify the performance gains achieved by training predictive models on augmented data. Additionally, expanding the generation pipeline to cover other specialized medical domains could further validate the generalization of this framework.

\appendix

\section{Acknowledgements}
This study is supported by the Dirección General de Investigación e Innovación Tecnológica de la Comunidad de Madrid (Orden 3177/2024) through the I+D Technological activities program (TEC-2024/COM-224).

This research was supported by CIBER -Consorcio Centro de Investigación Biomédica en Red- (CB/07/09/0025), the Instituto de Salud Carlos III with the support of the European Regional Development Fund (ISCIII PI23/00614; PMP24/00026), by Fundació la Marató de TV3 (202226-31) and by CaixaResearch Health 2023 LCF/PR/HR23/52430033.

\bibliographystyle{vancouver} 
\bibliography{refs}

@article{guo2025deepseek,
      title={DeepSeek-R1: Incentivizing Reasoning Capability in LLMs via Reinforcement Learning}, 
      author={DeepSeek-AI},
      year={2025},
      eprint={2501.12948},
      archivePrefix={arXiv},
      primaryClass={cs.CL},
      url={https://arxiv.org/abs/2501.12948},
      journal={arXiv preprint arXiv:2501.12948},
}

@article{gretton2012kernel,
  title={A kernel two-sample test},
  author={Gretton, Arthur and Borgwardt, Karsten M and Rasch, Malte J and Sch{\"o}lkopf, Bernhard and Smola, Alexander},
  journal={The journal of machine learning research},
  volume={13},
  number={1},
  pages={723--773},
  year={2012},
  publisher={JMLR. org}
}

@article{zhang2019bertscore,
  title={Bertscore: Evaluating text generation with bert},
  author={Zhang, Tianyi and Kishore, Varsha and Wu, Felix and Weinberger, Kilian Q and Artzi, Yoav},
  journal={arXiv preprint arXiv:1904.09675},
  year={2019}
}

@inproceedings{clark2019sentence,
  title={Sentence mover’s similarity: Automatic evaluation for multi-sentence texts},
  author={Clark, Elizabeth and Celikyilmaz, Asli and Smith, Noah A},
  booktitle={Proceedings of the 57th annual meeting of the association for computational linguistics},
  pages={2748--2760},
  year={2019}
}

@inproceedings{lin2004rouge,
  title={Rouge: A package for automatic evaluation of summaries},
  author={Lin, Chin-Yew},
  booktitle={Text summarization branches out},
  pages={74--81},
  year={2004}
}

@inproceedings{banerjee2005meteor,
  title={METEOR: An automatic metric for MT evaluation with improved correlation with human judgments},
  author={Banerjee, Satanjeev and Lavie, Alon},
  booktitle={Proceedings of the acl workshop on intrinsic and extrinsic evaluation measures for machine translation and/or summarization},
  pages={65--72},
  year={2005}
}

@inproceedings{zhu2018texygen,
  title={Texygen: A benchmarking platform for text generation models},
  author={Zhu, Yaoming and Lu, Sidi and Zheng, Lei and Guo, Jiaxian and Zhang, Weinan and Wang, Jun and Yu, Yong},
  booktitle={The 41st international ACM SIGIR conference on research \& development in information retrieval},
  pages={1097--1100},
  year={2018}
}

@book{templin1957certain,
  title={Certain language skills in children; their development and interrelationships.},
  author={Templin, Mildred C},
  year={1957},
  publisher={University of Minnesota Press},
}

@article{clark1954distance,
  title={Distance to nearest neighbor as a measure of spatial relationships in populations},
  author={Clark, Philip J and Evans, Francis C},
  journal={Ecology},
  volume={35},
  number={4},
  pages={445--453},
  year={1954},
  publisher={JSTOR}
}

@article{mcinnes2018umap,
  title={Umap: Uniform manifold approximation and projection for dimension reduction},
  author={McInnes, Leland and Healy, John and Melville, James},
  journal={arXiv preprint arXiv:1802.03426},
  year={2018}
}

@article{van2008visualizing,
  title={Visualizing data using t-SNE.},
  author={Van der Maaten, Laurens and Hinton, Geoffrey},
  journal={Journal of machine learning research},
  volume={9},
  number={11},
  year={2008}
}

@book{who2016icd10,
  title={International Statistical Classification of Diseases and Related Health Problems (ICD-10)},
  author={{World Health Organization}},
  volume={1},
  year={2016},
  edition={5th},
  publisher={World Health Organization},
  address={Geneva},
  url={https://icd.who.int/browse10/2016/en}
}

@misc{OpenBioLLMs,
  author = {Ankit Pal, Malaikannan Sankarasubbu},
  title = {OpenBioLLMs: Advancing Open-Source Large Language Models for Healthcare and Life Sciences},
  year = {2024},
  publisher = {Hugging Face},
  journal = {Hugging Face repository},
  howpublished = {\url{https://huggingface.co/aaditya/OpenBioLLM-Llama3-70B}}
}

@misc{qwen3.5,
    title  = {{Qwen3.5}: Towards Native Multimodal Agents},
    author = {{Qwen Team}},
    month  = {February},
    year   = {2026},
    url    = {https://qwen.ai/blog?id=qwen3.5}
}

@inproceedings{wei2019eda,
  title={{EDA}: Easy Data Augmentation Techniques for Boosting Performance on Text Classification Tasks},
  author={Wei, Jason and Zou, Kai},
  booktitle={Proceedings of EMNLP-IJCNLP},
  pages={6382--6388},
  year={2019}
}

@article{shorten2021text,
  title={Text data augmentation for deep learning},
  author={Shorten, Connor and Khoshgoftaar, Taghi M and Furht, Borko},
  journal={Journal of big Data},
  volume={8},
  number={1},
  pages={101},
  year={2021},
  publisher={Springer}
}

@article{spasic2020clinical,
  title={Clinical text data in machine learning: systematic review},
  author={Spasic, Irena and Nenadic, Goran},
  journal={JMIR medical informatics},
  volume={8},
  number={3},
  pages={e17984},
  year={2020},
  publisher={JMIR Publications Toronto, Canada}
}

@article{guo2021augmenting,
  title={Augmenting data with mixup for sentence classification: An empirical study},
  author={Guo, Hongyu and Mao, Yi and Zhang, Richong},
  journal={arXiv preprint arXiv:1905.08941},
  year={2021}
}

@article{chai2026text,
  title={Text data augmentation for large language models: A comprehensive survey of methods, challenges, and opportunities},
  author={Chai, Yaping and Xie, Haoran and Qin, Joe S},
  journal={Artificial Intelligence Review},
  volume={59},
  number={1},
  pages={35},
  year={2026},
  publisher={Springer}
}

@article{dai2025auggpt,
  title={Auggpt: Leveraging chatgpt for text data augmentation},
  author={Dai, Haixing and Liu, Zhengliang and Liao, Wenxiong and Huang, Xiaoke and Cao, Yihan and Wu, Zihao and Zhao, Lin and Xu, Shaochen and Zeng, Fang and Liu, Wei and others},
  journal={IEEE Transactions on Big Data},
  volume={11},
  number={3},
  pages={907--918},
  year={2025},
  publisher={IEEE}
}

@article{del2025comparative,
  title={Comparative analysis of generative LLMs for labeling entities in clinical notes},
  author={del Moral-Gonz{\'a}lez, Rodrigo and G{\'o}mez-Adorno, Helena and Ramos-Flores, Orlando},
  journal={Genomics \& Informatics},
  volume={23},
  number={1},
  pages={3},
  year={2025},
  publisher={Springer}
}

@article{lyu2023hybrid,
  title={A hybrid framework with large language models for rare disease phenotyping},
  author={Lyu, Qingyu and others},
  journal={Journal of Biomedical Informatics / PMC},
  year={2023}
}
\end{document}